\documentclass[10pt,conference,a4paper]{IEEEtran}
\usepackage{cmap} 
\usepackage[british]{babel}
\usepackage[utf8]{inputenc}
\usepackage{amsmath}
\usepackage[T1]{fontenc}
\usepackage{graphicx, multirow}
\usepackage{algorithm, algpseudocode}
\usepackage{txfonts}
\usepackage{units}
\usepackage{enumitem}
\usepackage{tabularx}
\usepackage[misc]{ifsym}
\usepackage{microtype} 
\usepackage{booktabs} 
\usepackage{xspace}
\usepackage{cite} 
\usepackage{fancyhdr}

\DeclareMathOperator{\tr}{tr} 
\DeclareMathOperator{\rank}{rank} 
\DeclareMathOperator{\diag}{diag} 
\DeclareMathOperator{\argmin}{argmin} 

\newcommand{\gC}{\ensuremath{C}} 
\newcommand{\gCRIT}{\ensuremath{\mathcal{J}}} 
\newcommand{\gCRITf}[1]{\ensuremath{\gCRIT\left(#1\right)}} 
\newcommand{\gD}{\ensuremath{D}} 
\newcommand{\gDELTA}{\ensuremath{\gB{\Delta}}} 
\newcommand{\gDELTAccH}[2]{\ensuremath{\gH{\delta}\left(#1,#2\right)}} 
\newcommand{\gG}{\ensuremath{\mathcal{G}}} 
\newcommand{\gGn}[1]{\ensuremath{\gB{g}_{#1}}} 
\newcommand{\gGnH}[1]{\ensuremath{\gH{\gB{g}}_{#1}}} 
\newcommand{\gGnc}[2]{\ensuremath{\gGn{#1}^{(#2)}}} 
\newcommand{\gGAMMAjt}[2]{\ensuremath{\gamma_{#1}\left(#2\right)}} 
\newcommand{\gI}{\ensuremath{\gB{I}}} 
\newcommand{\gIc}[1]{\ensuremath{\mathcal{I}_{#1}}} 
\newcommand{\gJ}{\ensuremath{J}} 
\newcommand{\gLAG}[2]{\ensuremath{\mathcal{L}\left(#1,#2\right)}} 
\newcommand{\gLAMBDA}{\ensuremath{\gB{\Lambda}}} 
\newcommand{\gLAMBDAd}[1]{\ensuremath{\lambda_{#1}}} 
\newcommand{\gLAMBDAn}[1]{\ensuremath{\ell_{#1}}} 
\newcommand{\gM}{\ensuremath{\mu}} 
\newcommand{\gMc}[1]{\ensuremath{\gM_{#1}}} 
\newcommand{\gMcH}[1]{\ensuremath{\gH{\gM}_{#1}}} 
\newcommand{\gN}{\ensuremath{N}} 
\newcommand{\gNc}[1]{\ensuremath{\gN_{#1}}} 
\newcommand{\gPHI}{\ensuremath{\gB{\Phi}}} 
\newcommand{\gPHId}[1]{\ensuremath{\gB{f}_{#1}}} 
\newcommand{\gPSI}{\ensuremath{\gB{\Psi}}} 
\newcommand{\gSIGMAc}[1]{\ensuremath{\gB{\Sigma}_{#1}}} 
\newcommand{\gSIGMAcH}[1]{\ensuremath{\gB{\gH{\Sigma}}_{#1}}} 
\newcommand{\gSIGMAb}{\ensuremath{\gSIGMAc{\mathrm{B}}}} 
\newcommand{\gSIGMAw}{\ensuremath{\gSIGMAc{\mathrm{W}}}} 
\newcommand{\gSIGMAt}{\ensuremath{\gSIGMAc{\mathrm{T}}}} 
\newcommand{\gSIGMAtH}{\ensuremath{\gSIGMAcH{\mathrm{T}}}} 
\newcommand{\gT}{\ensuremath{T}} 

\newcommand{\gTHETA}{\ensuremath{\gB{\Theta}}}
\newcommand{\gXI}{\ensuremath{\gB{\Xi}}}
\newcommand{\gUPSILON}{\ensuremath{\gB{\Upsilon}}}

\newcommand{\gCHI}{\ensuremath{\gB{X}}}

\newcommand{\gOMEGA}{\ensuremath{\gB{\Omega}}}

\newcommand{\gB}[1]{\ensuremath{\mathbf{#1}}} 
\newcommand{\gH}[1]{\ensuremath{\widehat{#1}}} 
\newcommand{\gO}[1]{\ensuremath{\overline{#1}}} 
\newcommand{\gL}[1]{\ensuremath{{#1}_L}} 
\newcommand{\gE}[1]{\ensuremath{{#1}_E}} 

\newcommand\etal{\textit{et al.}\xspace}
\bgroup
\catcode`_=\active
\gdef\underworks{\catcode`_=\active
\def_{\setbox0=\hbox{0}\hskip\wd0\relax}}
\egroup

\usepackage[breaklinks=true,colorlinks,bookmarks=false]{hyperref}

\begin{document}

\title{Learning~Robust~Features for~Gait~Recognition by~Maximum~Margin~Criterion}

\author{\IEEEauthorblockN{Michal Balazia (\href{https://orcid.org/0000-0001-7153-9984}{ORCID 0000-0001-7153-9984}) and Petr Sojka (\href{https://orcid.org/0000-0002-5768-4007}{ORCID 0000-0002-5768-4007})}\IEEEauthorblockA{Faculty of Informatics, Masaryk University, Botanick\'a 68a, 602\,00 Brno, Czech Republic\\{\tt xbalazia@mail.muni.cz} and {\tt sojka@fi.muni.cz}}}

\maketitle

\pagestyle{plain}
\thispagestyle{fancy}
\fancyhead[C]{23rd IEEE/IAPR International Conference on Pattern Recognition 2016, preprint}

\begin{abstract}
In the field of gait recognition from motion capture data, designing human-interpretable gait features is a common practice of many fellow researchers. To refrain from ad-hoc schemes and to find maximally discriminative features we may need to explore beyond the limits of human interpretability. This paper contributes to the state-of-the-art with a machine learning approach for extracting robust gait features directly from raw joint coordinates. The features are learned by a modification of Linear Discriminant Analysis with Maximum Margin Criterion so that the identities are maximally separated and, in combination with an appropriate classifier, used for gait recognition. Experiments on the CMU MoCap database show that this method outperforms eight other relevant methods in terms of the distribution of biometric templates in respective feature spaces expressed in four class separability coefficients. Additional experiments indicate that this method is a leading concept for rank-based classifier systems.
\end{abstract}

\section{Introduction}
\label{intro}

From the surveillance perspective, gait pattern biometrics is appealing for its possibility of being performed at a distance and without body-invasive equipment or subject's cooperation. This allows data acquisition without a subject's consent. As the data are collected with high participation rate and surveilled subjects are not expected to claim their identities, the trait is preferably employed for identification rather than for authentication.

\begin{figure}[ht]
\centering
\includegraphics[width=0.48\textwidth]{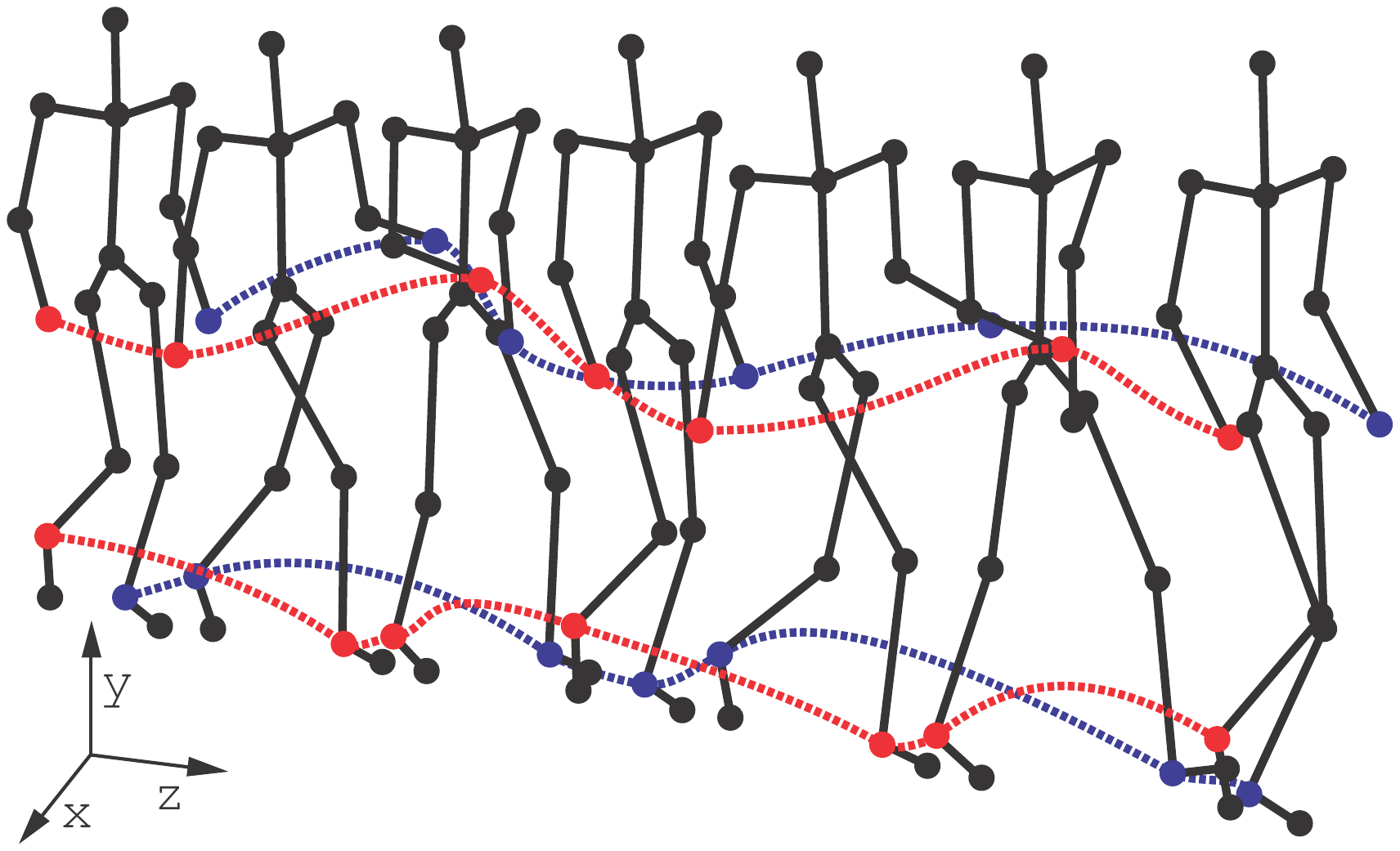}
\caption{Motion capture data. Skeleton is represented by a stick figure of 31~joints (only 17 are drawn here). Seven selected video frames of a walk sequence contain 3D coordinates of each joint in time. The red and blue lines track trajectories of hands and feet.~\cite{VBZ16}}
\label{f1}
\vspace{-16pt}
\end{figure}

Motion capture technology provides video clips of walking individuals containing structural motion data. The format keeps an overall structure of the human body and holds estimated 3D positions of major anatomical landmarks as the person moves. These so-called motion capture data (MoCap) can be collected online by a system of multiple cameras (Vicon) or a depth camera (Microsoft Kinect). To visualize motion capture data (see Figure~\ref{f1}), a simplified stick figure representing the human skeleton (a graph of joints connected by bones) can be recovered from the values of body point spatial coordinates. With recent rapid improvement in MoCap sensor accuracy, we believe in an affordable MoCap technology that can be installed in the streets and identify people from MoCap data.

Primary goal of this work is to project a method for learning robust gait features from raw MoCap data. A collection of extracted features builds a gait template that serves as the walker's signature. Templates are stored in a central database. Recognition of a person involves capturing their walk sample, extracting gait features to compose a template, and finally querying this database for a set of similar templates to report the most likely identity. Similarity of two templates is expressed in a single number computed by a similarity/distance function.

Many geometric gait features have been introduced over the past few years: Ahmed~\etal~\cite{AAS14} extract horizontal and vertical distances of selected joint pairs. Andersson~\etal~\cite{AA15} calculate mean and standard deviation in the signals of lower joint (hips, knees and ankles) angles. Ball~\etal~\cite{BRRV12} select mean, standard deviation and maximum of the signals of lower joint (hips, knees and ankles) angles. Dikovski~\etal~\cite{DMG14} construct 7~different feature sets from a broad spectrum of geometric features, such as static body parameters, joint angles and inter-joint distances aggregated within a gait cycle, along with various statistics. Kwolek~\etal~\cite{KKMJ14} extract bone rotations, inter-joint distances, and the person's height. Preis~\etal~\cite{PKWL12} define thirteen pose attributes, eleven of them static body parameters and the other two dynamic parameters: step length and walk speed. Sinha~\etal~\cite{SCB13} combine a number of gait features: areas of upper and lower body, inter-joint distances and all features introduced by Ball~\etal~\cite{BRRV12} and Preis~\etal~\cite{PKWL12}. Clearly, joint angles and step length are schematic and human-interpretable, which is convenient for visualizations and for intuitive understanding of the models, but unnecessary for automatic gait recognition. This application prefers learning features that maximally separate the identities and are not limited by such dispensable factors. Section~\ref{meth} gives a scheme for learning the features by Maximum Margin Criterion.

We have implemented and evaluated all methods on a testing database described in Section~\ref{exp-dat}. Their performance is expressed in several evaluation metrics defined in Section~\ref{exp-pm}. Results are presented and discussed in Section~\ref{exp-comp}.

\section{Learning Gait Features}
\label{meth}

In statistical pattern recognition, reducing space dimensionality is a common technique to overcome class estimation problems. Classes are discriminated by projecting high-dimensional input data onto low-dimensional sub-spaces by linear transformations with the goal of maximizing the class separability. We are interested in finding an optimal feature space where a gait template is close to those of the same walker and far from those of different walkers.

Let the model of a human body have $\gJ$ joints and all samples be linearly normalized to their average length~$\gT$. Labeled learning data in a sample (measurement) space have the form $\gL{\gG}=\left\{\left(\gGn{n},\gLAMBDAn{n}\right)\right\}_{n=1}^{\gL{\gN}}$ where
\begin{equation}
\gGn{n}=\left[[\gGAMMAjt{1}{1}\,\cdots\,\gGAMMAjt{\gJ}{1}]^\top\,\cdots\,[\gGAMMAjt{1}{\gT}\,\cdots\,\gGAMMAjt{\gJ}{\gT}]^\top\right]^\top
\end{equation}
is a gait sample (one gait cycle) in which $\gGAMMAjt{j}{t}\in\mathbb{R}^3$ are 3D spatial coordinates of a joint $j\in\left\{1,\ldots,\gJ\right\}$ at time $t\in\left\{1,\ldots,\gT\right\}$ normalized with respect to the person's position and walk direction. See that $\gL{\gG}$ has dimensionality $\gD=3\gJ\gT$. Each learning sample falls strictly into one of the learning identity classes $\left\{\gIc{c}\right\}_{c=1}^{\gC}$. A class $\gIc{c}$ has $\gNc{c}$ samples. Here $\gIc{c}\cap\gIc{c'}=\emptyset$ for $c \neq c'$ and $\gL{\gG}=\bigcup_{c=1}^{\gC}\gIc{c}$. $\gLAMBDAn{n}$ is the ground-truth label of the walker's identity class. We say that samples $\gGn{n}$ and $\gGn{n'}$ share a common walker if they are in the same class, i.e., $\gGn{n},\gGn{n'}\in\gIc{c}\Leftrightarrow\gLAMBDAn{n}=\gLAMBDAn{n'}$.

We measure class separability of a given feature space by a representation of the Maximum Margin Criterion (MMC) used by the Vapnik's Support Vector Machines (SVM)~\cite{V95}
\begin{equation}
\gCRIT=\frac{1}{2}\sum_{c,c'=1}^{\gC}\left(\left(\gMc{c}-\gMc{c'}\right)^\top\left(\gMc{c}-\gMc{c'}\right)-\tr\left(\gSIGMAc{c}+\gSIGMAc{c'}\right)\right)
\end{equation}
which is actually a summation of $\frac{1}{2}\gC(\gC-1)$ between-class margins. The margin is defined as the Euclidean distance of class means minus both individual variances (traces of scatter matrices $\gSIGMAc{c}=\frac{1}{\gNc{c}}\sum_{n=1}^{\gNc{c}}\left(\gGnc{n}{c}-\gMc{c}\right)\left(\gGnc{n}{c}-\gMc{c}\right)^\top$ and similarly for $\gSIGMAc{c'}$). For the whole labeled data, we denote the between- and within-class and total scatter matrices
\begin{equation}
\begin{split}
\gSIGMAb & =\sum_{c=1}^{\gL{\gC}}\left(\gMc{c}-\gM\right)\left(\gMc{c}-\gM\right)^\top\\
\gSIGMAw & =\sum_{c=1}^{\gL{\gC}}\frac{1}{\gNc{c}}\sum_{n=1}^{\gNc{c}}\left(\gGnc{n}{c}-\gMc{c}\right)\left(\gGnc{n}{c}-\gMc{c}\right)^\top\\
\gSIGMAt & =\sum_{c=1}^{\gL{\gC}}\frac{1}{\gNc{c}}\sum_{n=1}^{\gNc{c}}\left(\gGnc{n}{c}-\gM\right)\left(\gGnc{n}{c}-\gM\right)^\top=\gSIGMAb+\gSIGMAw
\end{split}
\end{equation}
where $\gGnc{n}{c}$ denotes the $n$-th sample in class $\gIc{c}$ and $\gMc{c}$ and $\gM$ are sample means for class $\gIc{c}$ and the whole data set, respectively, that is, $\gMc{c}=\frac{1}{\gNc{c}}\sum_{n=1}^{\gNc{c}}\gGnc{n}{c}$ and $\gM=\frac{1}{\gL{\gN}}\sum_{n=1}^{\gL{\gN}}\gGn{n}$. Now we obtain
\begin{equation}
\begin{split}
\gCRIT & \!=\!\frac{1}{2}\sum_{c,c'=1}^{\gC}\left(\gMc{c}-\gMc{c'}\right)^\top\left(\gMc{c}-\gMc{c'}\right)-\frac{1}{2}\sum_{c,c'=1}^{\gC}\tr\left(\gSIGMAc{c}+\gSIGMAc{c'}\right)\\
& \!=\!\frac{1}{2}\sum_{c,c'=1}^{\gC}\left(\gMc{c}-\gM+\gM-\gMc{c'}\right)^\top\left(\gMc{c}-\gM+\gM-\gMc{c'}\right)-\sum_{c=1}^{\gC}\tr\left(\gSIGMAc{c}\right)\\
& \!=\!\tr\left(\sum_{c=1}^{\gC}\left(\gMc{c}-\gM\right)\left(\gMc{c}-\gM\right)^\top\right)-\tr\left(\sum_{c=1}^{\gC}\gSIGMAc{c}\right)\\
& \!=\!\tr\left(\gSIGMAb\right)-\tr\left(\gSIGMAw\right)=\tr\left(\gSIGMAb-\gSIGMAw\right).
\end{split}
\end{equation}
Since $\tr\left(\gSIGMAb\right)$ measures the overall variance of the class mean vectors, a large one implies that the class mean vectors scatter in a large space. On the other hand, a small $\tr\left(\gSIGMAw\right)$ implies that classes have a small spread. Thus, a large $\gCRIT$ indicates that samples are close to each other if they share a common walker but are far from each other if they are performed by different walkers. Extracting features, that is, transforming the input data in the measurement space into a feature space of higher $\gCRIT$, can be used to link new observations of walkers more successfully.

Feature extraction is given by a linear transformation (feature) matrix $\gPHI\in\mathbb{R}^{\gD\times\gH{\gD}}$ from a $\gD$-dimensional measurement space $\gG=\left\{\gGn{n}\right\}_{n=1}^{\gN}$ of not necessarily labeled gait samples into a $\gH{\gD}$-dimensional feature space $\gH{\gG}=\left\{\gGnH{n}\right\}_{n=1}^{\gN}$ of gait templates where $\gH{\gD}<\gD$. Gait samples $\gGn{n}$ are transformed into gait templates $\gGnH{n}=\gPHI^\top\gGn{n}$. The objective is to learn a transform $\gPHI$ that maximizes the accumulated margin of the feature space
\begin{equation}
\gCRITf{\gPHI}=\tr\left(\gPHI^\top\left(\gSIGMAb-\gSIGMAw\right)\gPHI\right).
\label{e2}
\end{equation}
Once the transformation is found, all measured samples are transformed into templates (in the feature space) along with the class means and covariances. The templates are compared by the Mahalanobis distance function
\begin{equation}
\gDELTAccH{\gGnH{n}}{\gGnH{n'}}=\sqrt{\left(\gGnH{n}-\gGnH{n'}\right)^\top\gSIGMAtH^{-1}\left(\gGnH{n}-\gGnH{n'}\right)}.
\label{e3}
\end{equation}

Now we show that solution to the optimization problem in Equation~\eqref{e2} can be obtained by eigendecomposition of the matrix $\gSIGMAb-\gSIGMAw$. An important property to notice about the objective $\gCRITf{\gPHI}$ is that it is invariant w.r.t.\@ rescalings $\gPHI\rightarrow\alpha\gPHI$. Since it is a scalar itself, we can always choose $\gPHI=\gPHId{1}\|\cdots\|\gPHId{\gH{\gD}}$ such that $\gPHId{\gH{d}}^\top\gPHId{\gH{d}}=1$ and reduce the problem of maximizing $\gCRITf{\gPHI}$ into the constrained optimization problem
\begin{equation}
\begin{split}
\max & \quad\sum_{\gH{d}=1}^{\gH{\gD}}\gPHId{\gH{d}}^\top\left(\gSIGMAb-\gSIGMAw\right)\gPHId{\gH{d}}\\
\mathrm{subject\,to} & \quad\gPHId{\gH{d}}^\top\gPHId{\gH{d}}-1=0\qquad\forall\gH{d}=1,\ldots,\gH{\gD}.
\end{split}
\end{equation}
To solve the above optimization problem, let us consider the Lagrangian
\begin{equation}
\gLAG{\gPHId{\gH{d}}}{\gLAMBDAd{\gH{d}}}=\sum_{\gH{d}=1}^{\gH{\gD}}\gPHId{\gH{d}}^\top\left(\gSIGMAb-\gSIGMAw\right)\gPHId{\gH{d}}-\gLAMBDAd{\gH{d}}\left(\gPHId{\gH{d}}^\top\gPHId{\gH{d}}-1\right)
\end{equation}
with multipliers $\gLAMBDAd{\gH{d}}$. To find the maximum, we derive it with respect to $\gPHId{\gH{d}}$ and equate to zero
\begin{equation}
\frac{\partial\gLAG{\gPHId{\gH{d}}}{\gLAMBDAd{\gH{d}}}}{\partial\gPHId{\gH{d}}}=\left(\left(\gSIGMAb-\gSIGMAw\right)-\gLAMBDAd{\gH{d}}\gI\right)\gPHId{\gH{d}}=0
\end{equation}
which leads to
\begin{equation}
\left(\gSIGMAb-\gSIGMAw\right)\gPHId{\gH{d}}=\gLAMBDAd{\gH{d}}\gPHId{\gH{d}}
\end{equation}
where $\gLAMBDAd{\gH{d}}$ are the eigenvalues of $\gSIGMAb-\gSIGMAw$ and $\gPHId{\gH{d}}$ are the corresponding eigenvectors. Therefore,
\begin{equation}
\gCRITf{\gPHI}=\tr\left(\gPHI^\top\left(\gSIGMAb-\gSIGMAw\right)\gPHI\right)=\tr\left(\gPHI^\top\gLAMBDA\gPHI\right)=\tr\left(\gLAMBDA\right)
\end{equation}
is maximized when $\gLAMBDA=\diag\left(\gLAMBDAd{1},\ldots,\gLAMBDAd{\gH{\gD}}\right)$ has $\gH{\gD}$ largest eigenvalues and $\gPHI$ contains the corresponding leading eigenvectors.

In the following we discuss how to calculate the eigenvectors of $\gSIGMAb-\gSIGMAw$ and to determine an optimal dimensionality $\gH\gD$ of the feature space. First, we rewrite $\gSIGMAb-\gSIGMAw=2\gSIGMAb-\gSIGMAt$. Note that the null space of $\gSIGMAt$ is a subspace of that of $\gSIGMAb$ since the null space of $\gSIGMAt$ is the common null space of $\gSIGMAb$ and $\gSIGMAw$. Thus, we can simultaneously diagonalize $\gSIGMAb$ and $\gSIGMAt$ to some $\gDELTA$ and $\gI$
\begin{equation}
\begin{split}
\gPSI^\top\gSIGMAb\gPSI & =\gDELTA\\
\gPSI^\top\gSIGMAt\gPSI & =\gI
\end{split}
\end{equation}
with the $\gD\times\rank\left(\gSIGMAt\right)$ eigenvector matrix
\begin{equation}
\gPSI=\gOMEGA\gTHETA^{-\frac{1}{2}}\gXI
\end{equation}
where $\gOMEGA$ and $\gTHETA$ are the eigenvector and eigenvalue matrices of $\gSIGMAt$, respectively, and $\gXI$ is the eigenvector matrix of $\gTHETA^{-1/2}\gOMEGA^\top\gSIGMAb\gOMEGA\gTHETA^{-1/2}$. To calculate $\gPSI$, we use a fast two-step algorithm~\cite{LJZ06} in virtue of Singular Value Decomposition (SVD). SVD expresses a real $r \times s$ matrix $\gB{A}$ as a product $\gB{A}=\gB{U}\gB{D}\gB{V}^\top$ where $\gB{D}$ is a diagonal matrix with decreasing non-negative entries, and $\gB{U}$ and $\gB{V}$ are $r\times\min\left\{r,s\right\}$ and $s\times\min\left\{r,s\right\}$ eigenvector matrices of $\gB{A}\gB{A}^\top$ and $\gB{A}^\top\gB{A}$, respectively, and the non-vanishing entries of $\gB{D}$ are square roots of the non-zero corresponding eigenvalues of both $\gB{A}\gB{A}^\top$ and $\gB{A}^\top\gB{A}$. See that $\gSIGMAt$ and $\gSIGMAb$ can be expressed in the forms
\begin{equation}
\begin{split}
\gSIGMAt=&\enskip\gCHI\gCHI^\top\enskip\mathrm{where}\enskip\gCHI=\frac{1}{\sqrt{\gL{\gN}}}\left[\left(\gGn{1}-\gM\right)\cdots\left(\gGn{\gL{\gN}}-\gM\right)\right]\enskip\text{and}\\
\gSIGMAb=&\enskip\gUPSILON\gUPSILON^\top\enskip\text{where}\enskip\gUPSILON=\left[\left(\gMc{1}-\gM\right)\cdots\left(\gMc{\gL{\gC}}-\gM\right)\right],
\end{split}
\end{equation}
respectively. Hence, we can obtain the eigenvectors $\gOMEGA$ and the corresponding eigenvalues $\gTHETA$ of $\gSIGMAt$ through the SVD of $\gCHI$ and analogically $\gXI$ of $\gTHETA^{-1/2}\gOMEGA^\top\gSIGMAb\gOMEGA\gTHETA^{-1/2}$ through the SVD of $\gTHETA^{-1/2}\gOMEGA^\top\gUPSILON$. The columns of $\gPSI$ are clearly the eigenvectors of $2\gSIGMAb-\gSIGMAt$ with the corresponding eigenvalues $2\gDELTA-\gI$. Therefore, to constitute the transform $\gPHI$ by maximizing the MMC, we should choose the eigenvectors in $\gPSI$ that correspond to the eigenvalues of at least $\frac{1}{2}$ in $\gDELTA$. Note that $\gDELTA$ contains at most $\rank\left(\gSIGMAb\right)=\gC-1$ positive eigenvalues.

We found inspiration in the Fisher's Linear Discriminant Analysis (LDA)~\cite{F36} that uses the Fisher's criterion
\begin{equation}
\gCRITf{\gPHI_\mathrm{LDA}}=\tr\left(\frac{\gPHI_\mathrm{LDA}^\top\gSIGMAb\gPHI_\mathrm{LDA}}{\gPHI_\mathrm{LDA}^\top\gSIGMAw\gPHI_\mathrm{LDA}}\right).
\end{equation}
However, since the rank of $\gSIGMAw$ is at most $\gL{\gN}-\gC$, it is a~singular (non-invertible) matrix if $\gL{\gN}$ is less than $\gD+\gC$, or, analogously might be unstable if $\gL{\gN}\ll\gD$. Small sample size is a substantial difficulty as it is necessary to calculate $\gSIGMAw^{-1}$. To alleviate this, the measured data can be first projected to a lower dimensional space using Principal Component Analysis (PCA), resulting in a two-stage PCA+LDA feature extraction technique~\cite{BHK97}
\begin{equation}
\begin{split}
\gCRITf{\gPHI_\mathrm{PCA}}=&\tr\left(\gPHI_\mathrm{PCA}^\top\gSIGMAt\gPHI_\mathrm{PCA}\right)\\
\gCRITf{\gPHI_\mathrm{LDA}}=&\tr\left(\frac{\gPHI_\mathrm{LDA}^\top\gPHI_\mathrm{PCA}^\top\gSIGMAb\gPHI_\mathrm{PCA}\gPHI_\mathrm{LDA}}{\gPHI_\mathrm{LDA}^\top\gPHI_\mathrm{PCA}^\top\gSIGMAw\gPHI_\mathrm{PCA}\gPHI_\mathrm{LDA}}\right)
\end{split}
\end{equation}
and the final transform is $\gPHI=\gPHI_\mathrm{PCA}\gPHI_\mathrm{LDA}$. Given that there are $\gO{\gD}$ principal components, then regardless of the dimensionality $\gD$ there are at least $\gO{\gD}+1$ independent data points. Thus, if the $\gO{\gD}\times\gO{\gD}$ matrix $\gPHI_\mathrm{PCA}^\top\gSIGMAw\gPHI_\mathrm{PCA}$ is estimated from $\gL{\gN}-\gC$ independent observations and providing the $\gC\leq\gO{\gD}\leq\gL{\gN}-\gC$, we can always invert $\gPHI_\mathrm{PCA}^\top\gSIGMAw\gPHI_\mathrm{PCA}$ and this way obtain the LDA estimate. Note that this method is sub-optimal for multi-class problems~\cite{LDH01} as PCA keeps at most $\gL{\gN}-\gC$ principal components whereas at least $\gL{\gN}-1$ of them are necessary in order not to lose information. PCA+LDA in this form has been used for silhouette-based (2D) gait recognition by Su~\etal~\cite{SLC09} and is included in our experiments with MoCap (3D).

On given labeled learning data~$\gL{\gG}$, Algorithm~\ref{a1} and Algorithm~\ref{a2} provided below are efficient ways of learning the transforms $\gPHI$ for MMC and PCA+LDA, respectively.

\begin{algorithm}[H]
\caption{LearnTransformationMatrixMMC$\left(\gL{\gG}\right)$}
\label{a1}
\begin{algorithmic}[1]
  \State split $\gL{\gG}=\left\{\left(\gGn{n},\gLAMBDAn{n}\right)\right\}_{n=1}^{\gL{\gN}}$ into $\left\{\gIc{c}\right\}_{c=1}^{\gL{\gC}}$ of $\gNc{c}=\left|\gIc{c}\right|$ samples
  \State compute overall mean $\gM=\frac{1}{\gL{\gN}}\sum_{n=1}^{\gL{\gN}}\gGn{n}$ and individual class means $\gMc{c}=\frac{1}{\gNc{c}}\sum_{n=1}^{\gNc{c}}\gGnc{n}{c}$
  \State compute $\gSIGMAb=\sum_{c=1}^{\gL{\gC}}\left(\gMc{c}-\gM\right)\left(\gMc{c}-\gM\right)^\top$
  \State compute $\gCHI=\frac{1}{\sqrt{\gL{\gN}}}\left[\left(\gGn{1}-\gM\right)\cdots\left(\gGn{\gL{\gN}}-\gM\right)\right]$
  \State compute $\gUPSILON=\left[\left(\gMc{1}-\gM\right)\cdots\left(\gMc{\gL{\gC}}-\gM\right)\right]$
  \State compute eigenvectors $\gOMEGA$ and corresponding eigenvalues $\gTHETA$ \newline of $\gSIGMAt$ through SVD of $\gCHI$
  \State compute eigenvectors $\gXI$ of $\gTHETA^{\nicefrac{-1}{2}}\gOMEGA^\top\gSIGMAb\gOMEGA\gTHETA^{\nicefrac{-1}{2}}$ \newline through SVD of $\gTHETA^{\nicefrac{-1}{2}}\gOMEGA^\top\gUPSILON$
  \State compute eigenvectors $\gPSI=\gOMEGA\gTHETA^{\nicefrac{-1}{2}}\gXI$
  \State compute eigenvalues $\gDELTA=\gPSI^\top\gSIGMAb\gPSI$
  \State return transform $\gPHI$ as eigenvectors in $\gPSI$ \newline that correspond to the eigenvalues of at least $\nicefrac{1}{2}$ in $\gDELTA$
\end{algorithmic}
\end{algorithm}
\vspace{-7pt}%
\begin{algorithm}[H]
\caption{LearnTransformationMatrixPCALDA$\left(\gL{\gG}\right)$}
\label{a2}
\begin{algorithmic}[1]
  \State split $\gL{\gG}=\left\{\left(\gGn{n},\gLAMBDAn{n}\right)\right\}_{n=1}^{\gL{\gN}}$ into $\left\{\gIc{c}\right\}_{c=1}^{\gL{\gC}}$ of $\gNc{c}=\left|\gIc{c}\right|$ samples
  \State compute overall mean $\gM=\frac{1}{\gL{\gN}}\sum_{n=1}^{\gL{\gN}}\gGn{n}$ and individual class means $\gMc{c}=\frac{1}{\gNc{c}}\sum_{n=1}^{\gNc{c}}\gGnc{n}{c}$
  \State compute $\gSIGMAb=\sum_{c=1}^{\gL{\gC}}\left(\gMc{c}-\gM\right)\left(\gMc{c}-\gM\right)^\top$
  \State compute $\gSIGMAw=\sum_{c=1}^{\gL{\gC}}\frac{1}{\gNc{c}}\sum_{n=1}^{\gNc{c}}\left(\gGnc{n}{c}-\gMc{c}\right)\left(\gGnc{n}{c}-\gMc{c}\right)^\top$
  \State compute eigenvectors $\gPHI_\mathrm{PCA}$ of $\gSIGMAt=\gSIGMAb+\gSIGMAw$ \newline that correspond to $\gO{\gD}$ largest eigenvalues (we set $\gO{\gD}=\gL{\gC}$)
  \State compute eigenvectors $\gPHI_\mathrm{LDA}$ \newline of $(\gPHI_\mathrm{PCA}^\top\gSIGMAw\gPHI_\mathrm{PCA})^{-1}(\gPHI_\mathrm{PCA}^\top\gSIGMAb\gPHI_\mathrm{PCA})$
  \State return transform $\gPHI=\gPHI_\mathrm{PCA}\gPHI_\mathrm{LDA}$
\end{algorithmic}
\end{algorithm}

\section{Experimental Evaluation}
\label{exp}

\subsection{Database}
\label{exp-dat}

For the evaluation purposes we have extracted a large number of samples from the general MoCap database from CMU~\cite{CMU03} as a well-known and recognized database of structural human motion data. It contains numerous motion sequences, including a considerable number of gait sequences. Motions are recorded with an optical marker-based Vicon system. People wear a black jumpsuit and have 41~markers taped on. The tracking space of \unit[30]{m$^2$}, surrounded by 12~cameras of sampling rate of \unit[120]{Hz} in the height from 2 to 4~meters above ground, creates a video surveillance environment. Motion videos are triangulated to get highly accurate 3D data in the form of relative body point coordinates (with respect to the root joint) in each video frame and stored in the standard ASF/AMC data format. Each registered participant is assigned with their respective skeleton described in an ASF file. Motions in the AMC files store bone rotational data, which is interpreted as instructions about how the associated skeleton deforms over time.

To use the collected data in a fairly manner, a prototypical skeleton is constructed and used to represent bodies of all subjects, shrouding the unique skeleton parameters of individual walkers. Assuming that all walking subjects are physically identical disables a trivial skeleton check as a potentially unfair 100\% classifier. Moreover, this is a skeleton-robust solution as all bone rotational data are linked with a fixed skeleton. To obtain realistic parameters, it is calculated as the mean of all skeletons in the provided ASF files.

3D joint coordinates are calculated using bone rotational data and the prototypical skeleton. One cannot directly use raw values of joint coordinates, as they refer to absolute positions in the tracking space, and not all potential methods are invariant to person's position or walk direction. To ensure such invariance, the center of the coordinate system is moved to the position of root joint $\gGAMMAjt{\mathrm{root}}{t}=[0,0,0]^\top$ for each time~$t$ and axes are adjusted to the walker's perspective: the X~axis is from right (negative) to left (positive), the Y~axis is from down (negative) to up (positive), and the Z~axis is from back (negative) to front (positive). In the AMC file structure notation it is achieved by zeroing the root translation and rotation (\texttt{root 0 0 0 0 0 0}) in all frames of all motion sequences.

Since the general motion database contains all motion types, we extracted a number of sub-motions that represent gait cycles. First, an exemplary gait cycle was identified, and clean gait cycles were then filtered out using the DTW distance over bone rotations. The similarity threshold was set high enough so that even the least similar sub-motion still semantically represents a gait cycle. Finally, subjects that contributed with less than 10~samples were excluded. The final database has 54~walking subjects that performed 3{,}843~samples in total, which makes an average of about 71~samples per subject.

\subsection{Performance Metrics}
\label{exp-pm}

\begin{figure}[t]
\centering
\includegraphics[width=0.48\textwidth]{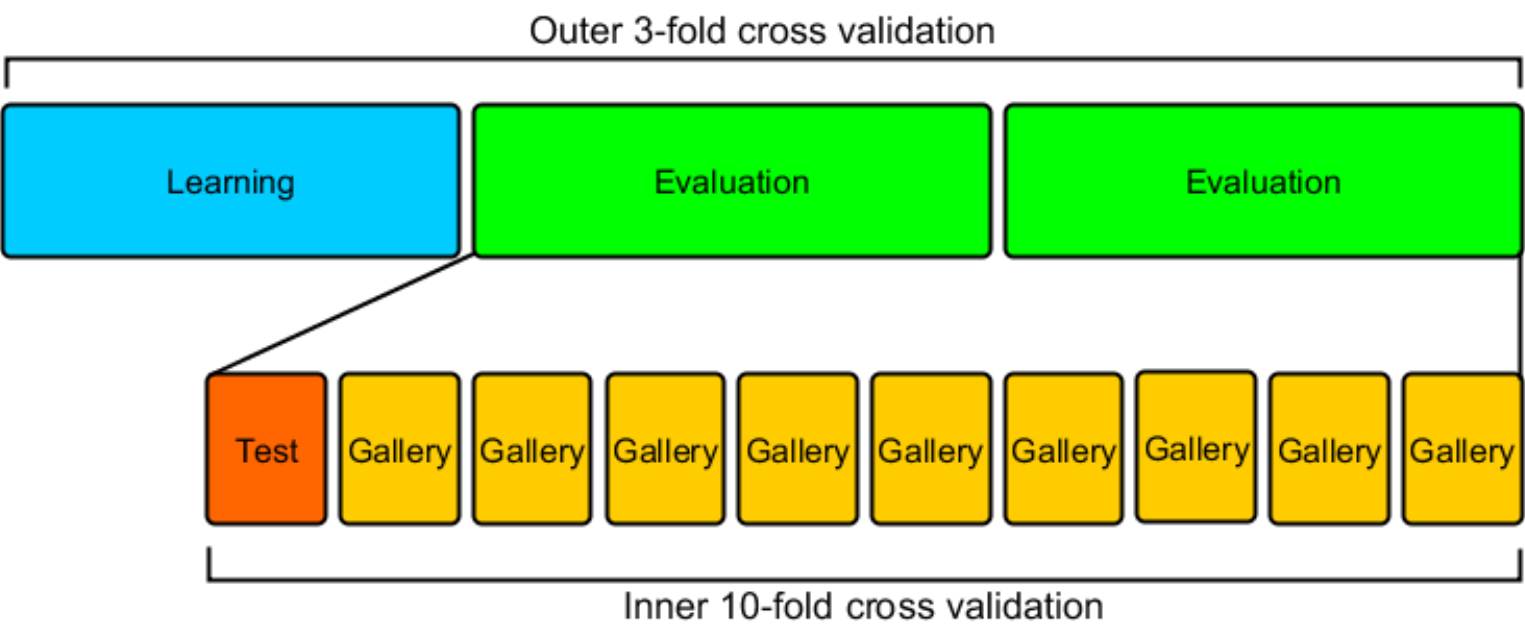}
\caption{Nested cross-validation.}
\label{f7}
\vspace{-5pt}
\end{figure}

All results are estimated with nested cross-validation (see Figure~\ref{f7}) that involves two partial cross-validation loops. In the outer 3-fold cross-validation loop, $\gL{\gN}$ labeled templates in one fold are used for learning the features and thus training the model. This model is frozen and ready to be evaluated for class separability coefficients on the remaining two folds of $\gE{\gN}$ labeled templates. Both learning and evaluation sets contain templates of all $\gC$ identities. Evaluation of classification metrics advances to the inner 10-fold cross-validation loop taking one dis-labeled fold as a testing set and other nine labeled folds as gallery. Test templates are classified by the winner-takes-all strategy, in which a test template $\gGnH{}^{\mathrm{test}}$ gets assigned with the label $\gLAMBDAn{\argmin_i\gDELTAccH{\gGnH{}^{\mathrm{test}}}{\gGnH{i}^{\mathrm{gallery}}}}$ of the gallery's closest identity class.

Correct Classification Rate (CCR) is often perceived as the ultimate qualitative measure; however, if a method has a low CCR, we cannot directly say if the system is failing because of bad features or a bad classifier. It is more explanatory to provide an evaluation in terms of class separability of the feature space. The class separability measures give an estimate on the recognition potential of the extracted features and do not reflect eventual combination with an unsuitable classifier:
\begin{description}[style=unboxed,leftmargin=0cm]
\def\ditem #1 (#2){\item[\textbullet\enskip\normalfont{\textit{#1:}}\enskip\textbf{#2}]}
\ditem Davies-Bouldin Index (DBI)
\begin{equation}
\mathrm{DBI}=\frac{1}{\gC}\sum_{c=1}^\gC\max\limits_{1 \leq c' \leq \gC,\,c' \neq c}\frac{\sigma_c+\sigma_{c'}}{\gDELTAccH{\gMcH{c}}{\gMcH{c'}}}
\end{equation}
where $\sigma_c=\frac{1}{\gNc{c}}\sum_{n=1}^{\gNc{c}}\gDELTAccH{\gGnH{n}}{\gMcH{c}}$ is the average distance of all elements in identity class $\gIc{c}$ to its centroid, and analogically for $\sigma_{c'}$. Templates of low intra-class distances and of high inter-class distances have a low DBI.
\ditem Dunn Index (DI)
\begin{equation}
\mathrm{DI}=\frac{\min\limits_{1 \leq c<c' \leq \gC}\gDELTAccH{\gMcH{c}}{\gMcH{c'}}}{\max\limits_{1 \leq c \leq \gC}\sigma_c}
\end{equation}
with $\sigma_c$ from the above DBI. Since this criterion seeks classes with high intra-class similarity and low inter-class similarity, a~high DI is more desirable.
\ditem Silhouette Coefficient (SC)
\begin{equation}
\mathrm{SC}=\frac{1}{\gE{\gN}}\sum_{n=1}^{\gE{\gN}}\frac{b(\gGnH{n})-a(\gGnH{n})}{\max\left\{a\left(\gGnH{n}),b(\gGnH{n}\right)\right\}}
\end{equation}
where $a(\gGnH{n})=\frac{1}{\gNc{c}}\sum_{n'=1}^{\gNc{c}}\gDELTAccH{\gGnH{n}}{\gGnH{n'}}$ is the average distance from $\gGnH{n}$ to other samples within the same identity class and $b(\gGnH{n})=\min\limits_{1 \leq c' \leq \gC,\,c' \neq c}\frac{1}{\gNc{c'}}\sum_{n'=1}^{\gNc{c'}}\gDELTAccH{\gGnH{n}}{\gGnH{n'}}$ is the average distance of $\gGnH{n}$ to the samples in the closest class. It is clear that $-1 \leq \mathrm{SC} \leq 1$ and SC close to one means that classes are appropriately separated.
\ditem Fisher's Discriminant Ratio (FDR)
\begin{equation}
\mathrm{FDR}=\frac{\frac{1}{\gC}\sum_{c=1}^\gC\gDELTAccH{\gMcH{c}}{\gH{\gM}}}{\frac{1}{\gE{\gN}}\sum_{c=1}^\gC\sum_{n=1}^{\gNc{c}}\gDELTAccH{\gGnH{n}}{\gMcH{c}}}.
\end{equation}
High FDR is preferred for seeking classes with low intra-class sparsity and high inter-class sparsity.
\end{description}

Apart from analyzing distribution of templates in the feature space, it is schematic to combine the features with a rank-based classifier and to evaluate the system based on distance distribution with respect to a probe. For obtaining a more applied performance evaluation, we evaluate:
\begin{description}[style=unboxed,leftmargin=0cm]
\def\ditem #1 (#2){\item[\textbullet\enskip\normalfont{\textit{#1:}}\enskip\textbf{#2}]~\newline}
\ditem Cumulative Match Characteristic (CMC) 
Sequence of Rank-$k$ (for $k$ on X axis from~1 up to the number of classes) recognition rates (Y~axis) for measuring ranking capabilities of a recognition method. Its headline Rank-1 is the well-known \textbf{CCR}.
\ditem False Accept Rate vs.\@ False Reject Rate (FAR/FRR)
Two sequences of the error rates (Y~axis) as functions of the discrimination threshold (X~axis). Each method has a value $e$ of this threshold giving Equal Error Rate (\textbf{EER}=FAR=FRR).
\ditem Receiver Operating Characteristic (ROC)
Sequence of True Accept Rate (TAR) and False Accept Rate with a varied discrimination threshold. For a given threshold the system signalizes both TAR (Y~axis) and FAR (X~axis). The value of Area Under Curve (\textbf{AUC}) is computed as the integral of the ROC curve.
\ditem Recall vs.\@ Precision (RCL/PCN)
Sequence of the rates with a~varied discrimination threshold. For a given threshold the system signalizes both RCL (X~axis) and PCN (Y~axis). The value of Mean Average Precision (\textbf{MAP}) is computed as the area under the RCL/PCN curve.
\end{description}

\subsection{Results of Comparative Evaluation}
\label{exp-comp}

\begin{figure*}[t]
\vspace{-5pt}
\centering
\begin{tabular}{@{}c@{ \ }c@{ }c@{}}
&&\multirow{1}{*}{\includegraphics[width=0.1\textwidth]{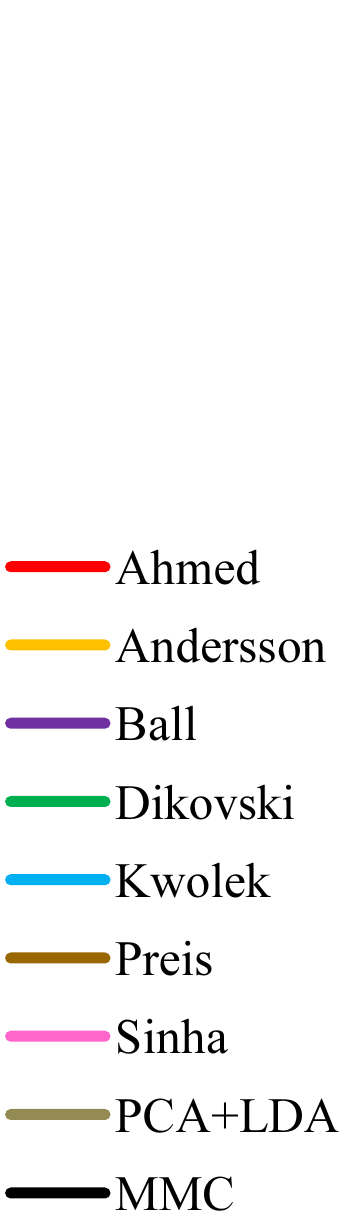}}\\
\includegraphics[width=0.44\textwidth]{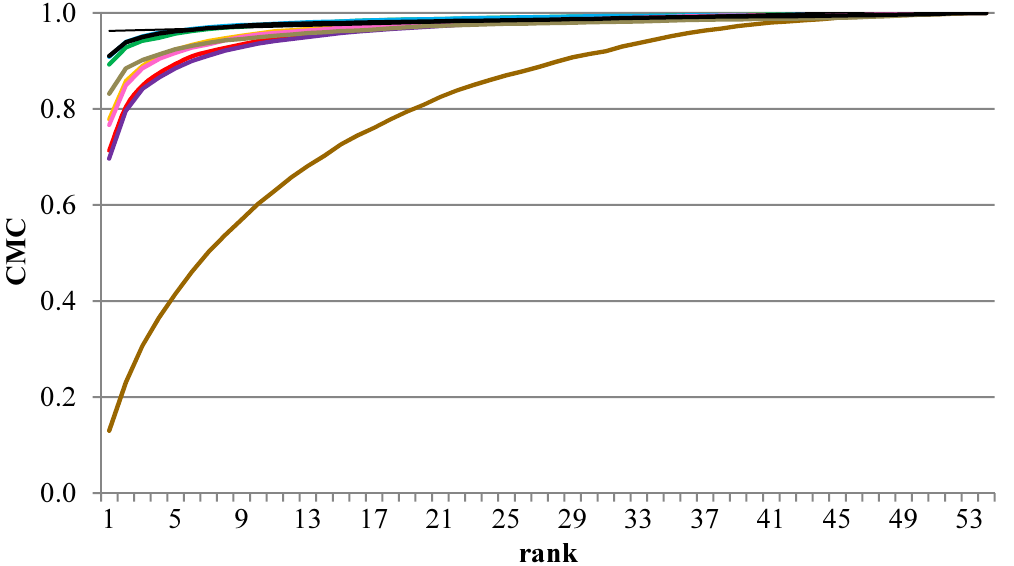}&\includegraphics[width=0.44\textwidth]{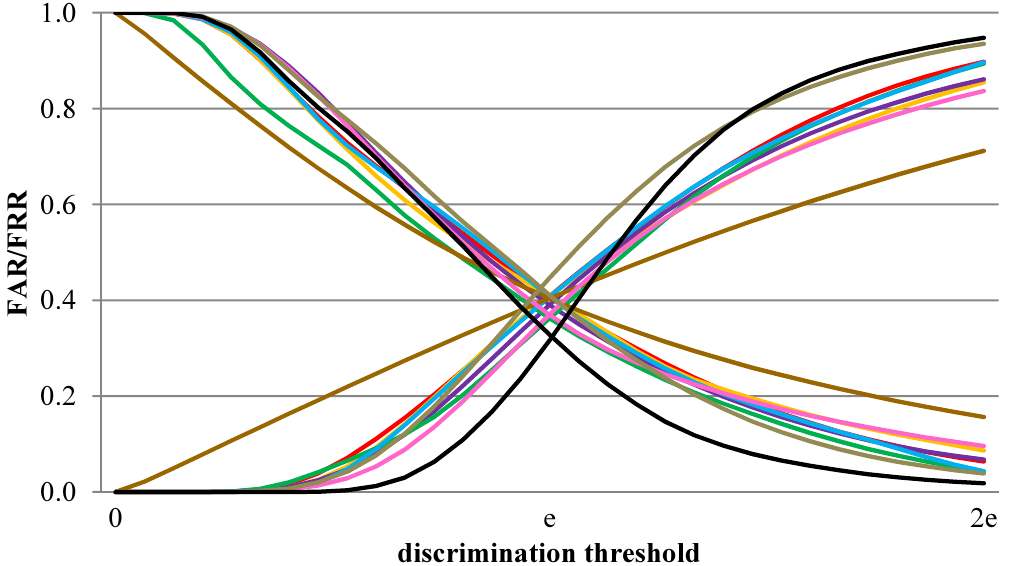}&\\
\includegraphics[width=0.44\textwidth]{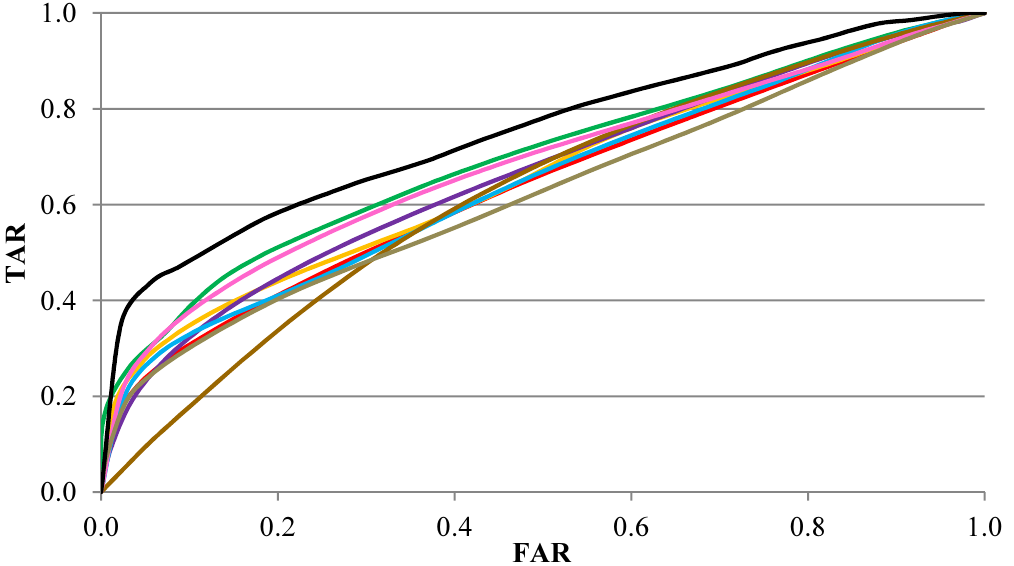}&\includegraphics[width=0.44\textwidth]{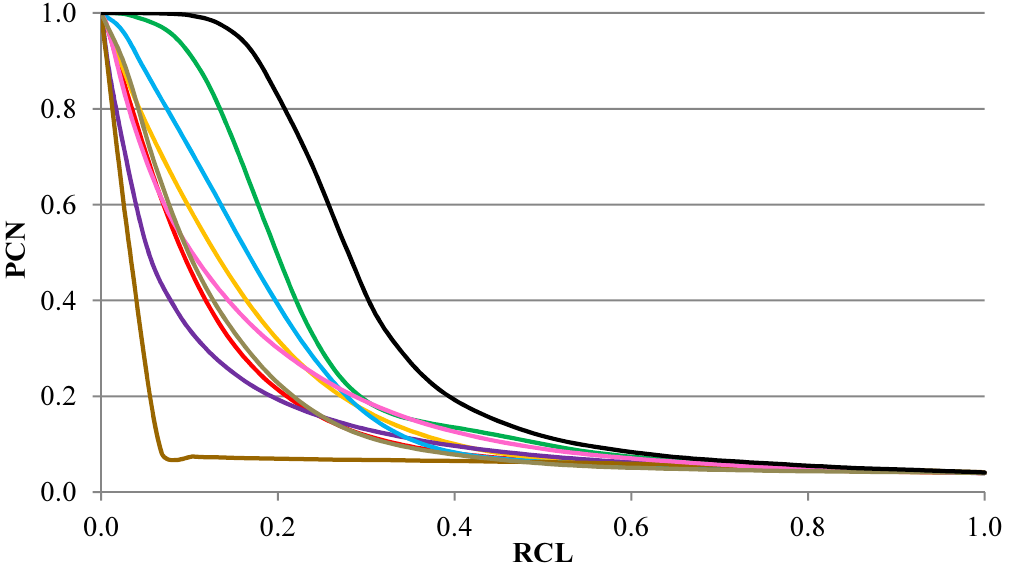}&
\end{tabular}
\vspace{-13pt}
\caption{Four rank-based classifier performance metrics CMC (top left), EER-centered FAR/FRR (top right), ROC (bottom left) and RCL/PCN (bottom right).}
\label{f5}
\vspace{-7pt}
\end{figure*}

In this section we provide comparative evaluation results of the proposed MMC-based feature extraction method against the related methods mentioned in Section~\ref{intro} in terms of performance metrics defined in Section~\ref{exp-pm}. All methods were implemented with their best-performance features, as Table~\ref{t1} shows. To ensure a fair comparison, we evaluate all methods on the same experimental database described in Section~\ref{exp-dat}. Figure~\ref{f5} and Table~\ref{t2} present results of the four class separability coefficients and the rank-based classification metrics.

\begin{table}[b]
\centering\underworks\tabcolsep8.4pt
\begin{tabular}{r|ll}
\toprule[1pt]
method & features & template dimensionality\\
\midrule[0.4pt]
Ahmed~\cite{AAS14} & HDF + VDF & 24\\
Andersson~\cite{AA15} & all described features & 80\\
Ball~\cite{BRRV12} & all described features & 18\\
Dikovski~\cite{DMG14} & Dataset 3 & 71\\
Kwolek~\cite{KKMJ14} & \texttt{g\_all} & 660\\
Preis~\cite{PKWL12} & static and dynamic & 13\\
Sinha~\cite{SCB13} & all described features & 45\\
PCA+LDA & learned by PCA+LDA & between $\gC$ and $\gL{\gN}-\gC$\\
MMC & learned by MMC & up to $\gC-1$\\
\bottomrule[1pt]
\end{tabular}
\vspace{3pt}
\caption{Configuration details of each implemented method. We advise readers to read the original papers for better understanding of the terminology in this table.}
\label{t1}
\end{table}

\begin{table*}[t]
\centering\underworks
\begin{tabular}{r@{\quad}|llll@{\qquad}|@{\qquad}llll}
\toprule[1pt]
& \multicolumn{4}{c|@{\qquad}}{class separability coefficients} & \multicolumn{4}{@{}c}{classification based metrics} \\
method					& _DBI				& _DI				& _SC							& _FDR				& CCR				& EER				& AUC					& MAP			\\
\midrule[0.4pt]
Ahmed~\cite{AAS14}		& _189.31			& 1.1823			& $-$0.1323						& 0.9213			& 0.7134			& 0.411				& 0.6387			& 0.1617			\\
Andersson~\cite{AA15}	& _175.5			& 1.2047			& $-$0.1117						& 0.9277			& 0.7787			& 0.413				& 0.6545			& 0.1926			\\
Ball~\cite{BRRV12}		& _200.75			& 1.014				& $-$0.14						& 0.982				& 0.6963			& 0.3906			& 0.6612			& 0.1454			\\
Dikovski~\cite{DMG14}	& _138.44			& 1.788				& $-$0.0873						& \textbf{1.2803}	& 0.8926			& 0.3625			& 0.6964			& 0.2582			\\
Kwolek~\cite{KKMJ14}	& _151.28			& 1.2027			& $-$0.074						& 0.974				& 0.9089			& 0.4072			& 0.6477			& 0.2121			\\
Preis~\cite{PKWL12}		& 2138.3			& 0.0313			& $-$0.3756						& 1.0957			& 0.13				& 0.4041			& 0.6236			& 0.0579			\\
Sinha~\cite{SCB13}		& _166.83			& 1.3				& $-$0.1256						& 1.0413			& 0.7666			& 0.3706			& 0.6809			& 0.1858			\\
\midrule[0.4pt]
PCA+LDA					& _195.19			& 1.021				& $-$0.084						& 0.8207			& 0.8314			& 0.447				& 0.6216			& 0.1643			\\
MMC						& \textbf{__96.2}	& \textbf{1.8453}	& \textbf{\hphantom{$-$}0.2227}	& 1.278				& \textbf{0.9102}	& \textbf{0.3223}	& \textbf{0.7551}	& \textbf{0.2996}	\\
\bottomrule[1pt]
\end{tabular}
\vspace{3pt}
\caption{Four class separability coefficients and supplementary rank-based classifier performance metrics related to Figure~\ref{f5} for all tested gait recognition methods.}
\label{t2}
\vspace{-27pt}
\end{table*}

The goal of the MMC-based learning is to find a linear discriminant that maximizes the misclassification margin. This optimization technique appears to be more effective than designing geometric gait features. Figure~\ref{f5} and in Table~\ref{t2} indicate the best results: lowest DBI, highest DI, highest (and exclusively positive) SC, second highest FDR and, combined with rank-based classifier, the best CMC, FAR/FRR, ROC and RCL/PCN scores along with all CCR, EER, AUC and MAP. We interpret the high scores as a sign of robustness. Apart from performance merits, the MMC method is also efficient: relatively low-dimensional templates (see the third column in Table~\ref{t1}) and Mahalanobis distance~\eqref{e3} ensure fast distance computations and thus contribute to high scalability.

\section{Conclusions and Future Work}
\label{conc}

The field of pattern recognition has recently advanced to an era where best results are often obtained using a machine learning approach. Finding optimal features for MoCap-based gait recognition is not an exception. This work introduces the concept of learning robust and discriminative features directly from raw joint coordinates by a modification of the Fisher's Linear Discriminant Analysis with Maximum Margin Criterion with the goal of maximal separation of identity classes. The introduced MMC method avoids instinctive drawing of ad-hoc features; on the contrary, they are computed from a much larger space beyond the limits of human interpretability. The collection of learned features achieves leading scores in four class separability coefficients and therefore has a strong potential in gait recognition applications. This is demonstrated by outperforming other methods in numerous rank-based classification metrics. We believe that MMC is a suitable criterion for optimizing gait features; however, our future work will continue with research on further potential optimality criterions and machine learning approaches.

Furthermore, we are investigating on whether these features can discriminate different people than exclusively who they are learned on. Can the number of learning identities be (much) smaller than the number of walkers encountered in the real operation? The main idea is to learn what aspects of walk people generally differ in and extract those as gait features. This is particularly important for a system to aid video surveillance applications where encountered walkers never supply labeled data and new identities can appear on the fly.

\medskip
\subsubsection*{Acknowledgments}
The data used in this project was created with funding from NSF EIA-0196217 and was obtained from \url{http://mocap.cs.cmu.edu}~\cite{CMU03}. Our extracted database and implementation details of the compared methods are available at \url{https://gait.fi.muni.cz} to support reproducibility of results.

\bibliographystyle{IEEEtran}
\bibliography{ref}

\end{document}